\documentclass[graybox]{svmult}
\usepackage{mathptmx}       % selects Times Roman as basic font
\usepackage{helvet}         % selects Helvetica as sans-serif font
\usepackage{courier}        % selects Courier as typewriter font
\usepackage{type1cm}        % activate if the above 3 fonts are                           
\usepackage{subfigure}
\usepackage{makeidx}         % allows index generation
\usepackage{graphicx}        % standard LaTeX graphics tool
\usepackage{multicol}        % used for the two-column index
\usepackage[bottom]{footmisc}% places footnotes at page bottom
\usepackage[colorlinks]{hyperref}
\usepackage{color,xcolor}
\usepackage{caption}
\usepackage{multirow}
\usepackage{indentfirst}
\usepackage{amsmath}

\DeclareGraphicsExtensions{.png,.jpg,.eps,.pdf}

\makeindex             % used for the subject index
\begin{document}

\title*{A Collaborative Aerial-Ground Robotic System for Fast Exploration}
% Use \titlerunning{Short Title} for an abbreviated version of
% your contribution title if the original one is too long
\author{Luqi Wang* \textsuperscript{1}, Daqian Cheng* \textsuperscript{2}, Fei Gao \textsuperscript{1}, Fengyu Cai \textsuperscript{1}, Jixin Guo \textsuperscript{1}, Mengxiang Lin \textsuperscript{2}, Shaojie Shen \textsuperscript{1}}
\authorrunning{L. Wang, D. Cheng, F. Gao, F. Cai, J. Guo, M. Lin, S. Shen.}
\institute{\textsuperscript{1} The Hong Kong University of Science and Technology \textsuperscript{2} Beihang University \\ **L. Wang and D. Cheng contributed equally to this work. \\
{\tt\small lwangax@connect.ust.hk, chengdaqian@buaa.edu.cn, fgaoaa@connect.ust.hk, fcaiaa@connect.ust.hk, jguoaj@connect.ust.hk, linmx@buaa.edu.cn, eeshaojie@connect.ust.hk}\\
This work was supported by HKUST project R9341.
}

%\author{Luqi Wang*, Daqian Cheng*, Fei Gao, Fengyu Cai, Jixin Guo, Mengxiang Lin, Shaojie Shen}
% Use \authorrunning{Short Title} for an abbreviated version of
% your contribution title if the original one is too long

%\institute{Luqi Wang \at Beihang University, Address of Institute, \email{chengdaqian@buaa.edu.cn}
%\and Name of Second Author \at Name, Address of Institute \email{name@email.address}}
%
% Use the package "url.sty" to avoid
% problems with special characters
% used your e-mail or web address
%
\maketitle

\vspace{-3cm}
\vspace{-0.3cm}

\begin{figure}[t]
\captionsetup{justification=centering}
\centering
{\includegraphics[width=0.95\columnwidth]{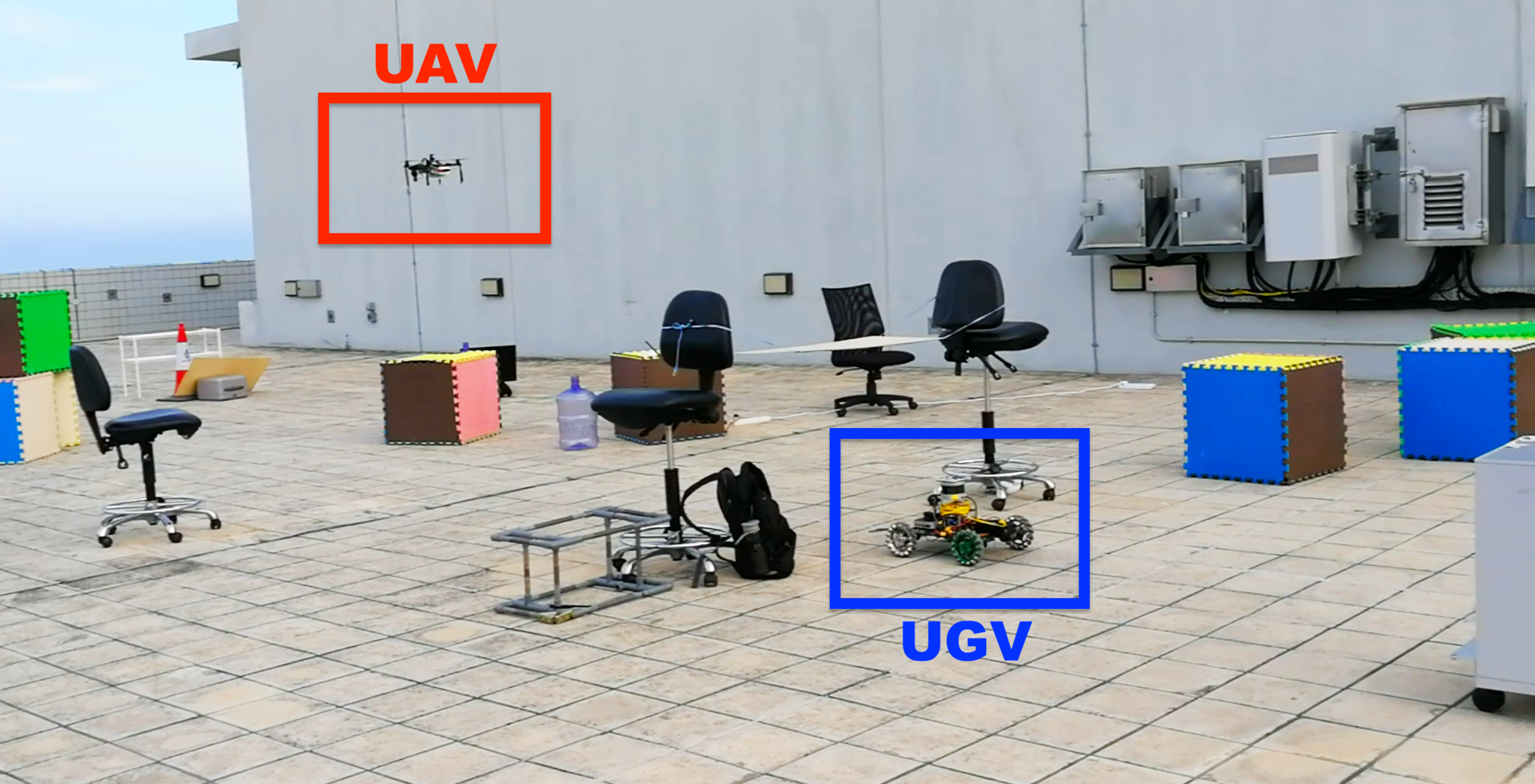}}
\caption{
\label{fig:experiment_site}  Outdoor exploration in progress.}
\vspace{-0.8cm}
\end{figure}

\section{Introduction}
\label{sec:motivation}
\vspace{-0.5cm}

%\subsection{Motivation and Problem Statement}

Exploration of unknown environments using autonomous robots has been considered a fundamental problem in robotics applications such as search and rescue~\cite{Shen2017Collaborative}, industrial inspection and 3D modelling. For exploration, the basic requirement for robots is to scan an unknown space or detect free space as fast as possible. Unmanned ground vehicles (UGVs)~\cite{yamauchi1997frontier} and unmanned aerial vehicles (UAVs)~\cite{cieslewski2017frontier} have both been employed for such tasks, with each having different advantages: UGVs are more payload-capable, so that they can carry heavy, long-range laser scanners which are inapplicable for weight-constrained UAVs. Meanwhile, UAVs have superiors mobility and agility, since they can fly above obstacles and cover areas that are inaccessible to UGVs, like obstacles' top surfaces. Consequently, a UGV often enjoys a larger sensor coverage, yet cluttered and view-blocking environments hamper its performance. On the other hand, a UAV may deliver inferior exploration efficiency due to its short-range sensor, but enjoys an unblocked downward-looking view. Therefore, UGVs are more favorable for open areas, while UAVs are preferable in cluttered environments. 

%\subsection{Related Works}
There are numerous works conducted on robotic exploration.~\cite{yamauchi1997frontier} first proposed the concept of frontiers, which are defined as unknown grid-map cells that are adjacent to free ones and thus represent accessible new information. Harmonic function, the solution to $Laplace's$ $Equation$, shown as Eq.~\ref{eq:laplace_equation}, is used to plan the path to frontiers~\cite{silva2002harmonic}. This method generates a scalar field in free-space based on its surrounding boundary conditions (occupied cells and frontier cells) and obtains the path using gradient-descent. These works are only conducted on ground platforms. For aerial-ground exploration, ~\cite{butzke2015airground} deploys a UAV as a back-up instead of as an independent explorer, only when the UGV encounters high, invisible areas. The system in~\cite{delmerico2016exploration} is also proposed based on the same idea that one vehicle helps another, failing to exploit both vehicles' full potential. 

In this paper, considering the different environmental advantages of UAVs and UGVs, we propose an autonomous collaborative framework which utilizes their complementary characteristics to achieve higher efficiency and robustness in exploration applications. We summarize our contribution as follows:
%\vspace{-0.3cm}
\begin{enumerate}
\item{An efficient exploration framework that combines UAVs' and UGVs' advantages.}
\item{A more efficient computation method for the harmonic function for robotic exploration tasks. }
\item{Integration of the proposed collaborative exploration framework with state estimation, sensor fusion and trajectory optimization modules. Extensive field experiments and simulations are presented to validate the efficiency and robustness of the proposed method.
}
\end{enumerate}
\vspace{-0.6cm}

\section{Environment Representation}
\label{sec:env_form}
\vspace{-0.4cm}

The information about the exploration environment is represented in a global grid map. Besides obstacles and free space, other information is required to be represented in the map, including frontiers and harmonic fields, to facilitate the exploration.

\vspace{-0.8cm}

\subsection{Frontier}
\label{subsec:frontier}
\vspace{-0.5cm}

The original definition of a frontier, the boundary that separates unknown space and free space, is proposed in~\cite{yamauchi1997frontier}. This definition of a frontier can be applied to UGVs directly. However, it is not suitable to directly apply to UAVs in some cases as they hover above obstacles. To facilitate the aerial-ground collaborative case, we define two types of frontier, as shown in Fig.~\ref{fig:sim_capture}:

\begin{enumerate}

\item Frontier A: The conventional definition of frontier proposed by~\cite{yamauchi1997frontier}, which is the boundary between free space and unknown space. This kind of frontier is visible to both ground and aerial vehicles.

\item Frontier B: A new type of frontier, which is the boundary between obstacles and unknown space. This kind of frontier usually occurs on the top surfaces of obstacles, and can, therefore, only be seen by an aerial vehicle.

\end{enumerate}

According to the advantages introduced stated in Sec.~\ref{sec:motivation}, we can intuitively conclude that UGVs pursue longer Frontier A lines, while UAVs are more favorable for Frontier B over Frontier A. The exploration planner is designed based on this intuition. In practice, we are using a grip map to store map information. Hence, the frontiers and other objects are all formed by grid cells in the exploration map.

\vspace{-0.5cm}

\subsection{Harmonic Field}
\label{subsec:harmonic_field}
\vspace{-0.5cm}

To facilitate the UGV path planning, we employ a harmonic function, formulating the exploration space as a potential field. The governing equation of potential $\phi$ in 2D is

\begin{equation}
\label{eq:laplace_equation}
\bigtriangledown^{2} \phi =\frac{\partial^{2} \phi }{\partial x^{2}}+\frac{\partial^{2} \phi }{\partial y^{2}}=0.
\end{equation}

The boundary conditions are set as below:
\begin{enumerate}
\item Occupied cells and the ground vehicle: zero potential value.
\item Frontier cells: the negative value of its frontier lines' lengths.
\end{enumerate}

\vspace{-0.5cm}

\section{Path Planning}
\label{sec:path_planning}

\vspace{-0.5cm}

To set up an effective exploration planner, we first formulate the proposed exploration-with-preference problem using a frontier method. Then a harmonic field method is adopted to generate a path towards one of the frontiers for the ground vehicle. Meanwhile, motion primitives are generated for evaluation and the primitive with the highest score is chosen as the target of the aerial vehicle. In the case that no primitive has a positive score, a target on the frontiers with the highest score is chosen. Finally, the paths are optimized and executed by the vehicles.

\vspace{-0.7cm}

\subsection{Terrestrial Planning}
\label{subsec:UGV_planning}

\vspace{-0.5cm}

\begin{figure}[t]
\centering
\subfigure[\label{fig:original_map} original map]
{\includegraphics[width=0.19\columnwidth]{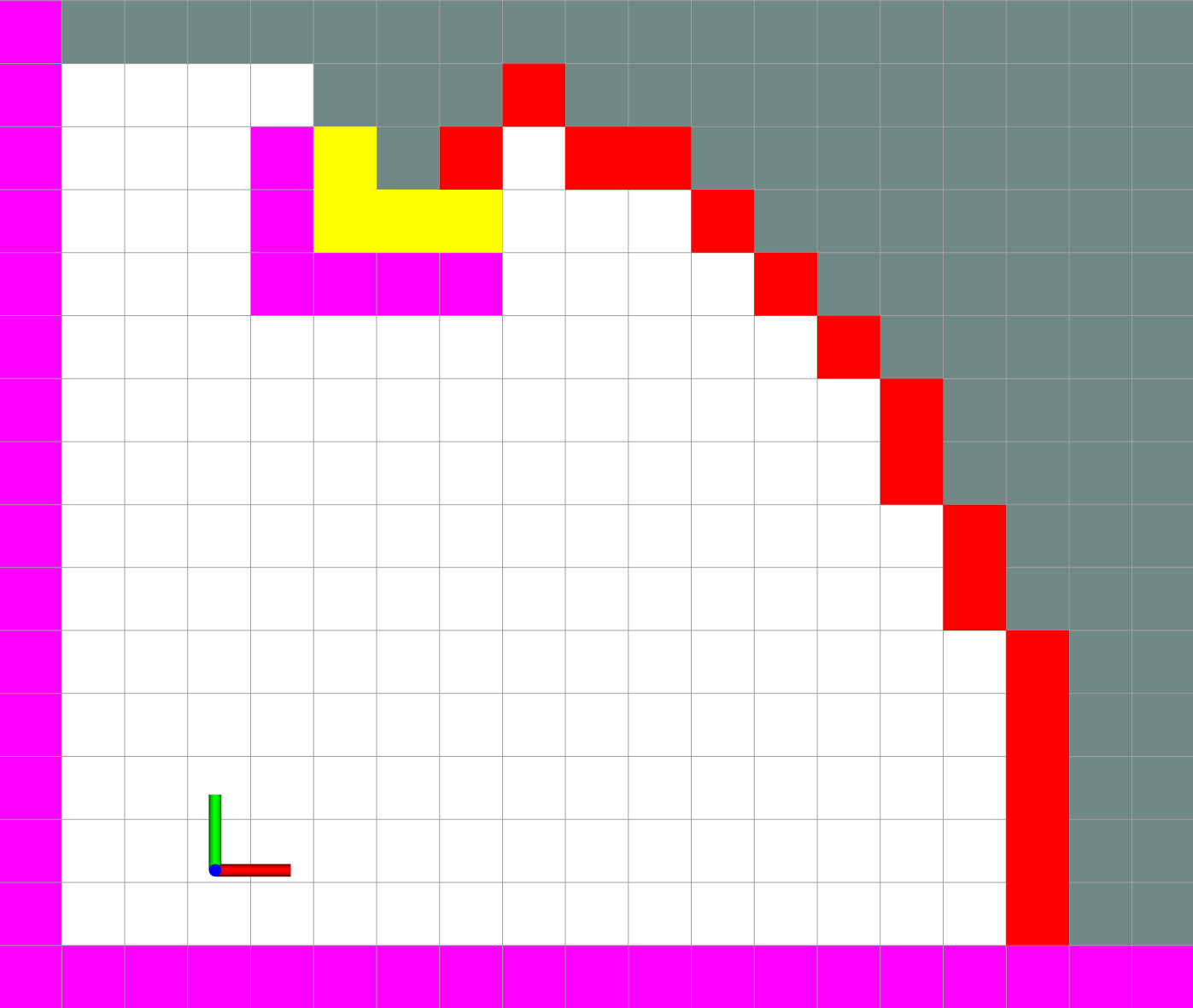}}
\subfigure[\label{fig:updated_map} updated map]
{\includegraphics[width=0.19\columnwidth]{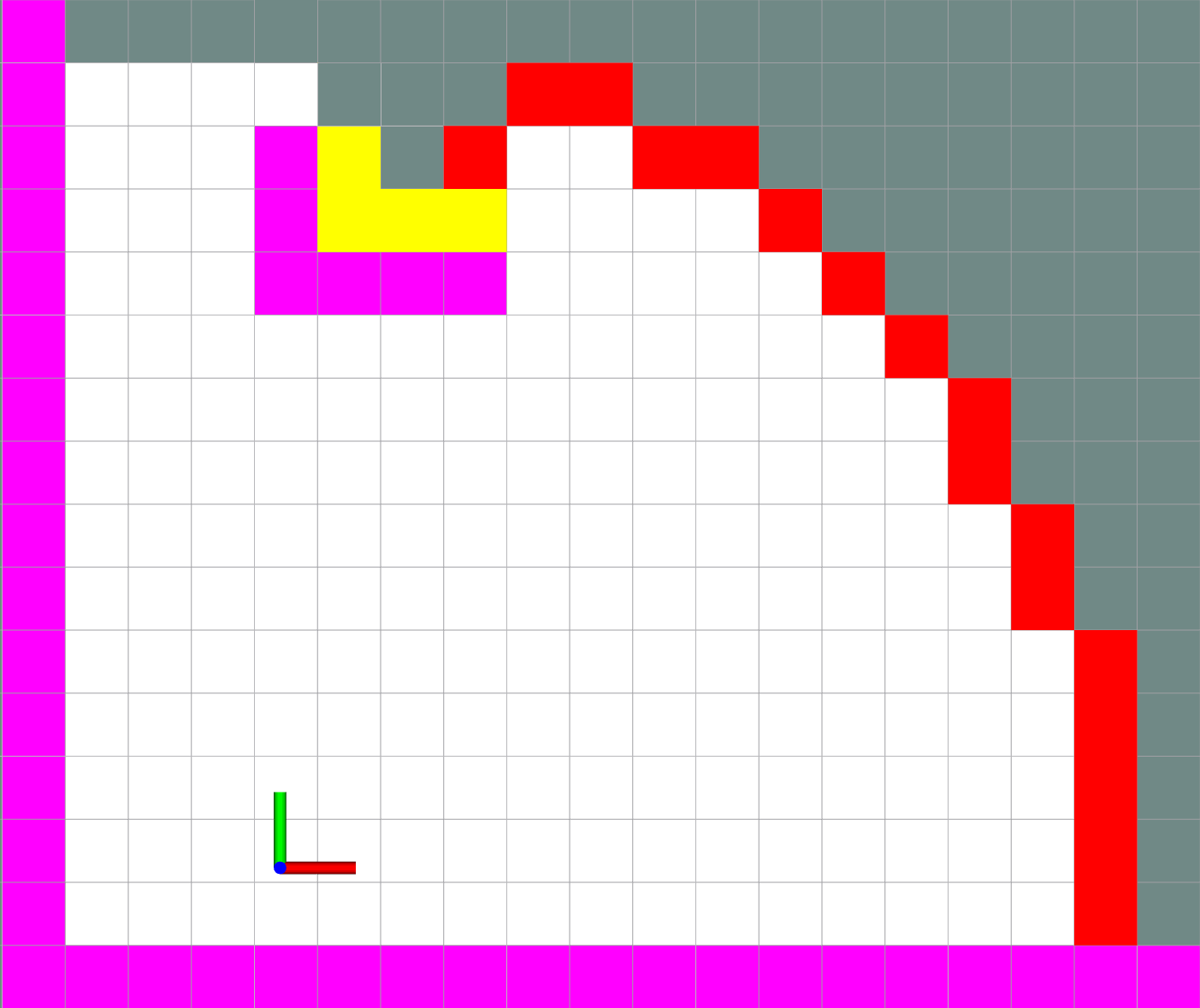}}
\subfigure[\label{fig:new_boundary} new boundary]
{\includegraphics[width=0.19\columnwidth]{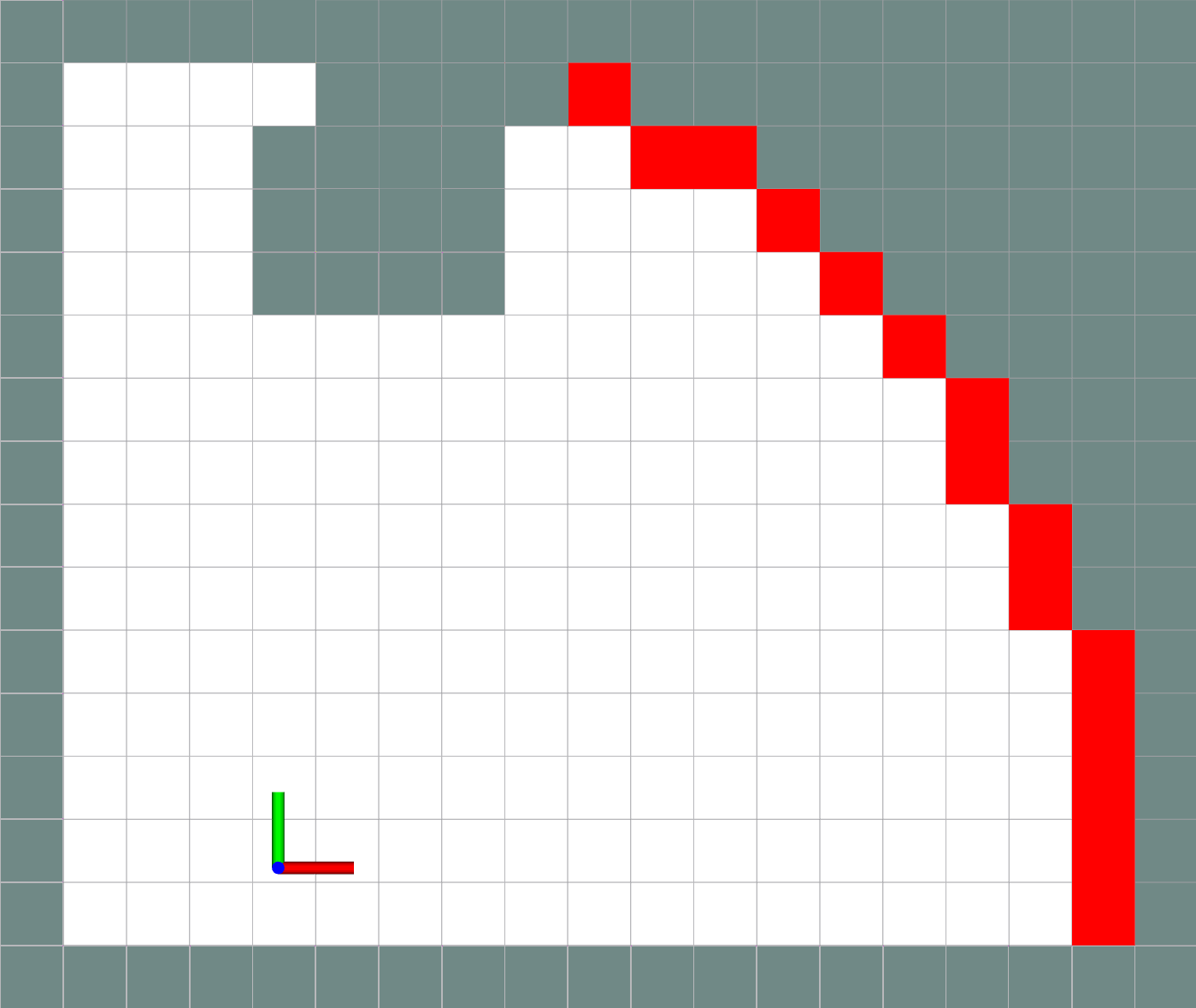}}
\subfigure[\label{fig:original_update_sequence} conventional sequence]
{\includegraphics[width=0.19\columnwidth]{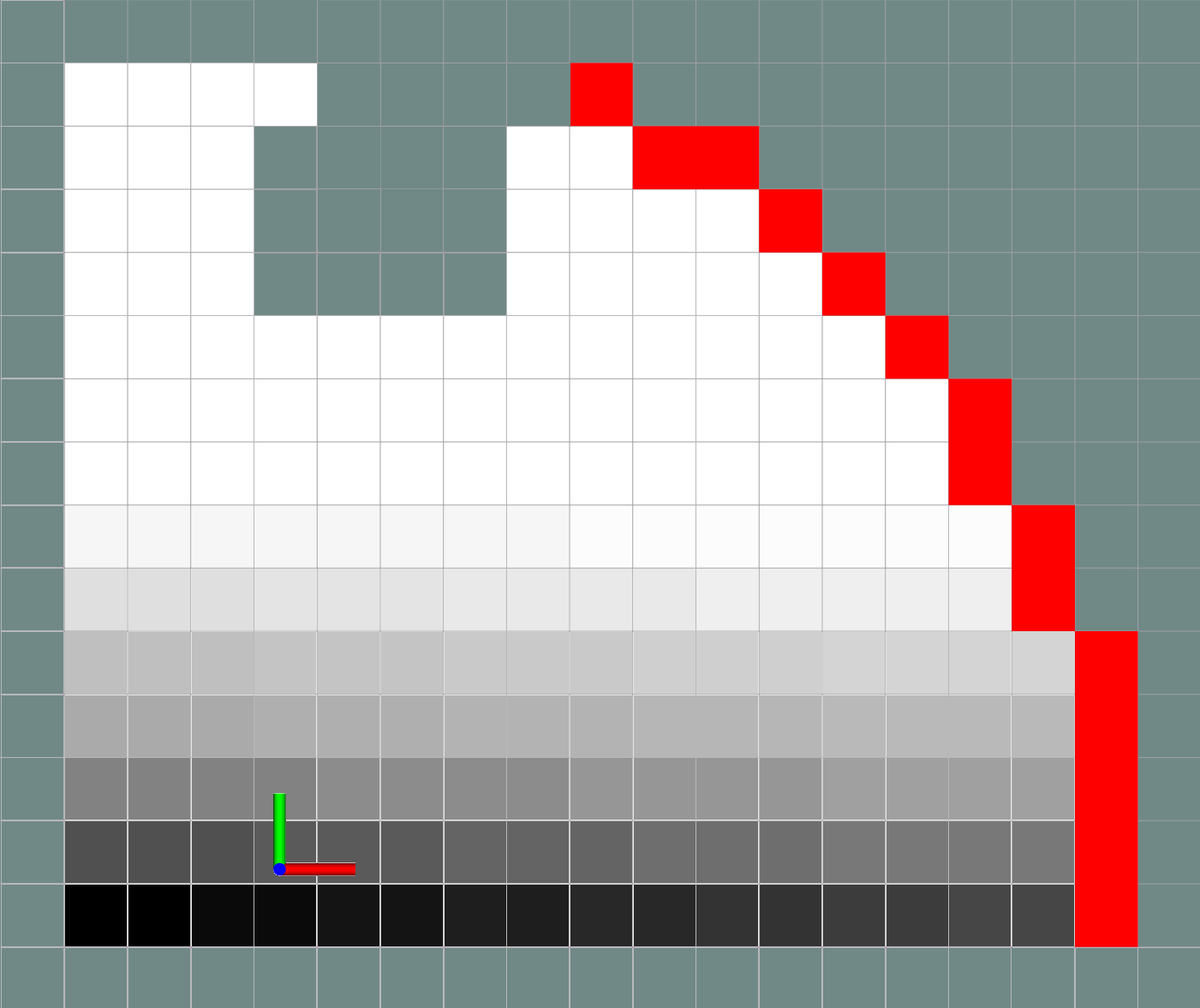}}
\subfigure[\label{fig:proposed_update_sequence} spreading sequence]
{\includegraphics[width=0.19\columnwidth]{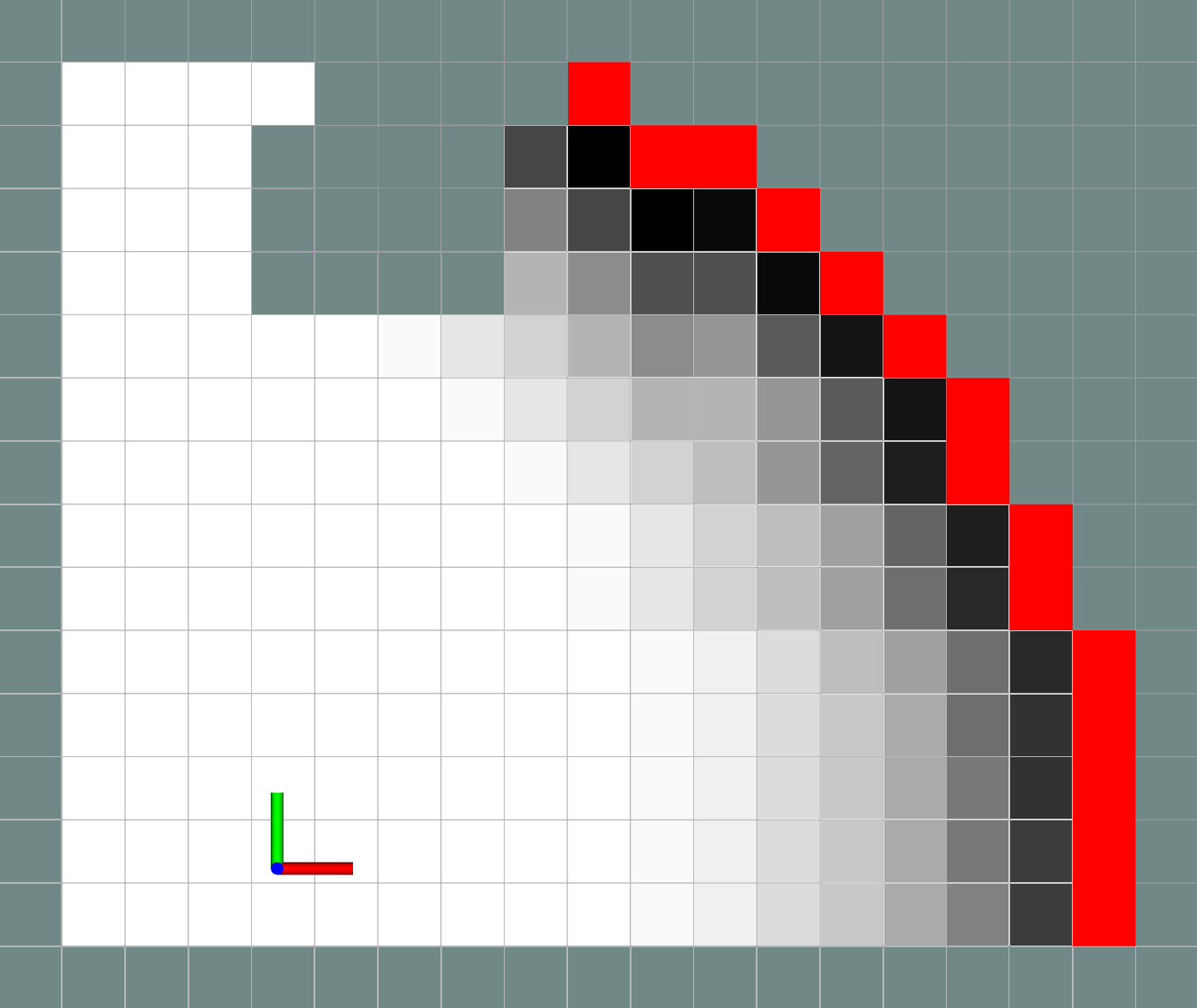}}
\vspace{-0.3cm}
\caption{ Illustration of harmonic field's update sequence. As the UGV moves forward, the map updates and new boundary cells are generated. In (c), white cells are free and need updating, gray cells are non-free and fixed-value, and red cells are new boundary cells. (d) and (e) show the conventional-sequence and spreading-sequence respectively; each cell's darkness represents its update sequence.
\label{fig:spreading_illustration}}
\vspace{-0.5cm}
\end{figure}

To utilize the formulated harmonic field mentioned in~\ref{subsec:harmonic_field} for UGV path planning, We first discretize the governing equation, Eq.~\ref{eq:laplace_equation}, and adopt the Successive over-relaxation (SOR) method~\cite{connolly1993harmonic} with relaxation parameter $w$, as shown in Eq.~\ref{eq:SOR}, to iteratively compute the potential value until convergence, and in each iteration, cells' potential values are updated sequentially. Such an update-sequence is conventionally the map-array's sequence (row-by-row), but we propose an update-sequence that spreads out from the new boundary conditions as shown in Fig.~\ref{fig:proposed_update_sequence}. This sequence is able to spread their influence faster, and the computation can converge within fewer iterations.

\begin{equation}
\label{eq:SOR}
\phi^{k+1}_{x,y}= (1-w)\phi^{k}_{x,y} + \frac{w}{4}(\phi^{k}_{x+1,y} + \phi^{k}_{x-1,y} + \phi^{k}_{x,y+1} + \phi^{k}_{x,y-1})
\end{equation}

When the field converges, a simple gradient-descent path can be obtained from the location of the ground vehicle.

Fig.~\ref{fig:sim_capture} gives an example of a harmonic function field and its gradient field, and the path obtained in this method has the following desirable features:
\begin{enumerate}
\item Being repelled from obstacles; thus the sensor's range is more utilized.
\item Being inclined to lead to longer and nearer frontier lines.
\end{enumerate}

\vspace{-0.5cm}

\subsection{Aerial Planning}
\label{subsec:UAV_planning}

\vspace{-0.5cm}

To achieve rapid exploration for the UAV, we employ a motion primitive method~\cite{likhachev2009planning} to generate path candidates for evaluation. The state space model is modeled as Eq.~\ref{eq:state_space} and the input is the body linear acceleration $a_{body}$:

\begin{equation}
\label{eq:state_space}
\dot{x}=Ax+Ba_{body}.
\end{equation}

With the state space model, the state transition equation can be derived to calculate the primitives:
\begin{equation}
x(t)=e^{At}x_0+\int_0^te^{A(t-\sigma)}Ba_{body}d\sigma.
\end{equation}

The evaluation of the a primitive $P$ is based on its score, which is formulated as the information gain along the path to the primitive divided by the cost to the primitive:

\begin{equation}
\label{eq:score}
P.Score = \frac{P.InfoGain}{P.Cost}.
\end{equation}

The cost is modeled as a linear quadratic minimum-time problem~\cite{verriest1991linear}:

\begin{equation}
\label{eq:cost}
P.Cost=\int_0^t (a_{body}^TRa_{body}+\rho) d\tau,
\end{equation}
where $\rho$ is the weight parameter tuning time and input and $R$ is the weight parameter of the acceleration in different directions. Larger $\rho$ implies a larger weight of time.

The information gain of a grid cell $C$ is modeled as

\begin{equation}
\label{eq:UAV_info_gain}
C.InfoGain = \sum_{f_i}^{Fv_{A}^{UAV}} w_A + \sum_{f_j}^{Fv_{B}^{UAV}} w_B + \sum_{u_k}^{Uv^{UAV}} w_U,
\end{equation}
where ${Fv_{A}^{UAV}}$, ${Fv_{B}^{UAV}}$ and ${Uv^{UAV}}$ are the set of visible Frontier A cells, Frontier B cells and unknown cells. As Frontier B cells indicate the top surface of the obstacles, Frontier B cell clusters indicate obstacle-cluttered areas, preferable for the aerial vehicle. Therefore, a larger weight should be assigned to Frontier B cells. In practice, $w_B$, $w_A$ and $w_U$ are set to 0.89, 0.1 and 0.01 respectively. The information gain along the path to the primitive is calculated generally by the summation of the information gain of the cells along the path, except that the visible cells are not counted repetitively.

\begin{figure}[t]
\center
\includegraphics[width=1.0\columnwidth]{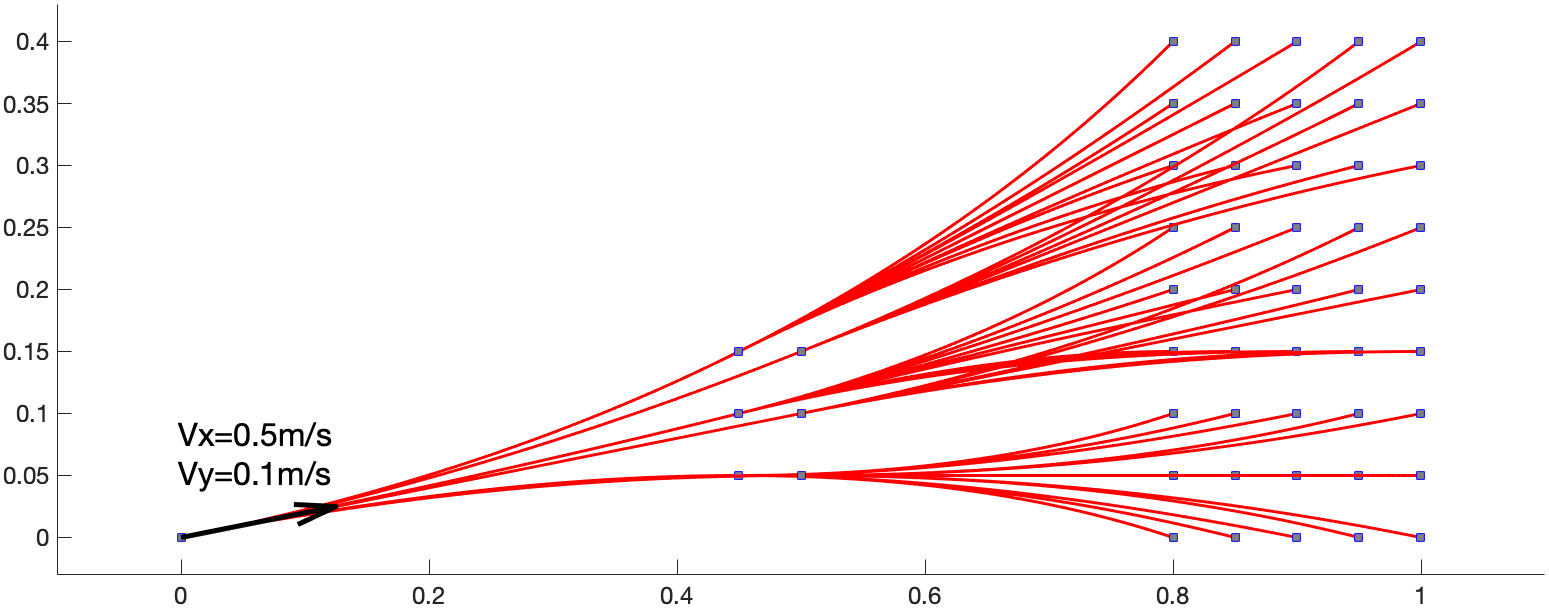}
\caption{Illustration of motion primitives with discretization of linear acceleration. The arrow at the initial position represents the initial velocity of 0.5m/s and 0.1m/s in x and y directions. The squares indicate the primitives, while the red curve represents the path to the primitives. Some of the primitives are trimmed since they exceed the speed limit of 0.5m/s on any of the axes. 
\label{fig:motion_primitive}}
\vspace{-0.5cm}
\end{figure}

An extensive primitive-tree is acquired by iteratively enforcing a discretized set of accelerations, as shown in Fig.~\ref{fig:motion_primitive}. Meanwhile, the primitives that exceed the speed limit are trimmed to ensure dynamic feasibility. Finally, the path candidate with the highest score is selected.

Nevertheless, restricted by the field of view (FOV) and the dynamic state of the aerial vehicle, it is possible that the paths to the primitives do not possess positive information gains or scores, since the planning strategy stated earlier in this section is just locally optimal. In this situation, a global target selector is adopted. The candidate target set is assigned to be all the frontier cells and the information gain at a cell is computed utilizing Eq.~\ref{eq:UAV_info_gain}. The costs to each of the frontier cells are set to be the Euclidean norm between the UAV's position and the frontier cells. The scores are calculated in the same way as mentioned in Eq.~\ref{eq:score} and the position with the highest score is selected as the target.

\vspace{-0.7cm}

\subsection{Trajectory Optimization}
\label{traj_opt}

\vspace{-0.5cm}

With the target path or position selected for UGV and UAV in Sec.~\ref{subsec:UGV_planning} and~\ref{subsec:UAV_planning}, the trajectories along the paths or towards the positions are required to be optimized against the energy cost while ensuring dynamic consistency for the vehicles to follow.

For the ground vehicle, we adopt the method described in our previous work~\cite{fei2018icra} for optimization. With a gradient-descent path generated, an obstacle-free travel corridor along it can be produced. Then, with the waypoints, continuity, safety, and dynamically feasibility constraints enforced, the energy cost, which is set to be the squared jerk of the trajectory, is optimized inside the corridor to generate piece-wise polynomial trajectories using the Bernstein polynomial basis. The $i^{th}$-order Bernstein polynomial basis is defined as

\begin{equation}
b_{n}^{i}(t) = \binom{n}{i} \cdot t^i \cdot (1-t)^{n-i},
\end{equation}
where $n$ is the degree of the Bernstein polynomial basis, $\binom{n}{i}$ is the binomial coefficient and $t$ is the variable parameterizing the trajectory. The waypoints constraints are enforced by assigning the initial and final positions, velocities and accelerations; the continuity constraints restrict the continuity of the derivatives between the curve pieces; the safety constraints add additional boundary limits to the corridor to ensure safety; and the dynamical feasibility constraints bound the velocity and the acceleration along the trajectories. 

For the aerial vehicle, we adopt a minimum jerk trajectory generator~\cite{mellinger2011minimum} to minimize the squared jerk with waypoint and continuity constraints enforced, which is similar to the ground vehicle, using polynomial trajectories.

\vspace{-0.5cm}

\section{System Description}
\label{sec:implentation_details}

\vspace{-0.5cm}

\begin{figure}[t]
\centering
\subfigure[\label{fig:UAV} Aerial platform]
{\includegraphics[width=0.48\columnwidth]{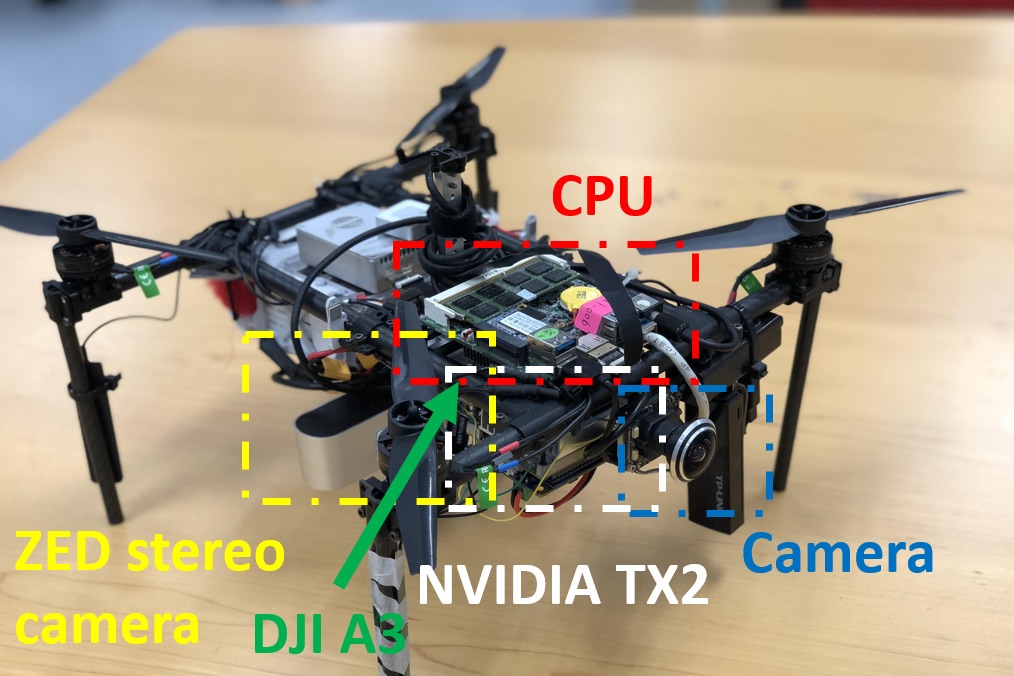}}
\subfigure[\label{fig:UGV} Ground platform]
{\includegraphics[width=0.48\columnwidth]{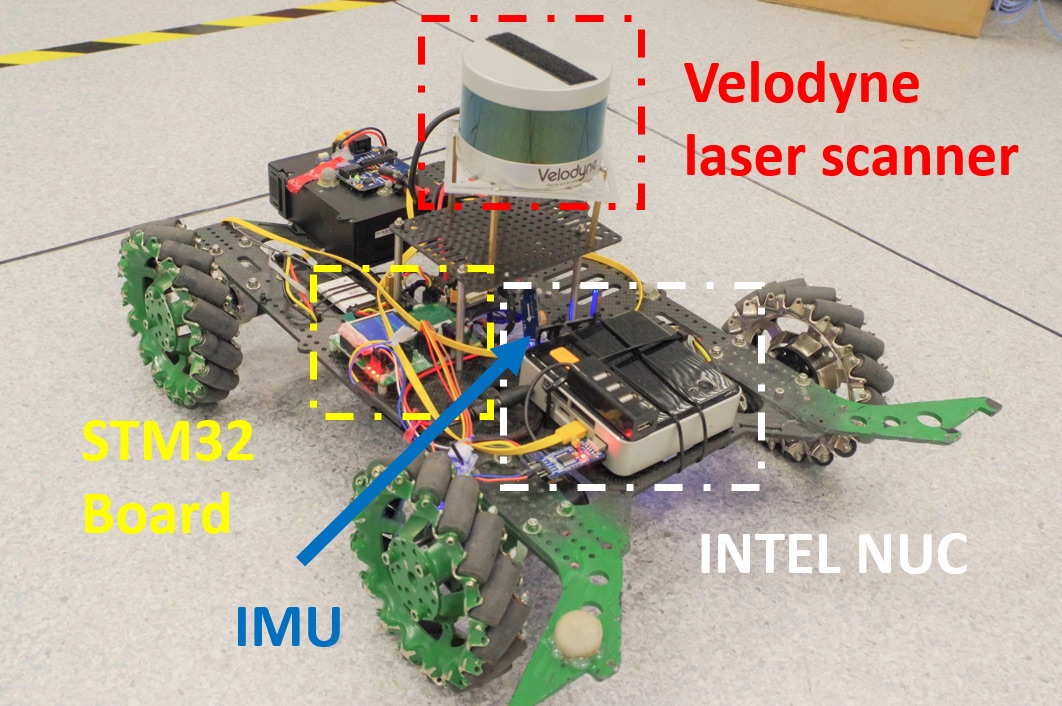}}
\vspace{-0.3cm}
\centering
\caption{ Experiment platforms.
\label{fig:system_description}}
\vspace{-0.7cm}
\end{figure}

The collaborative system contains an aerial platform and a ground platform. Both of the platforms shown in Fig.~\ref{fig:system_description} are custom-built vehicles, which can localize themselves and perform mapping in dense and complex environments. The exploration planner collects the real-time dynamic states of the vehicles to plan trajectories for them to follow. There is a wireless network set up for the communication between the vehicles.

\vspace{-0.7cm}

\subsection{Ground Platform}
\label{ground_platform}

\vspace{-0.5cm}

Our ground platform is an omni-directional vehicle with four Mecanum wheels, driven by four DJI RM 3510 brushless motor, which are installed on a carbon fiber chassis. A Velodyne VLP-16 laser scanner with a range up to 100 m is mounted above the chassis to serve both localization with an inertial measurement unit (IMU) and mapping purposes. The laser scan data are collected by an on-board Intel NUC with an i7-5500U processor, running Ubuntu 16.04 and ROS Kinetic, through Ethernet. The NUC is also responsible for other computation, including state estimation, mapping and high-level control. The velocity commands are sent to a STM32F407 microcontroller board through a serial port to directly control the motors on the chassis.

Localization, mapping and control programs are implemented on the NUC. An Extended Kalman Filter (EKF) is implemented to fuse IMU data and odometry data from the SLAM program LOAM~\cite{loam} for real-time localization. The scanned laser point cloud is collected and transformed into the world frame to integrate into the globally maintained map. Meanwhile, the fused odometries are sent to the exploration planner to generate trajectories for the vehicle to follow. The pose controller receives trajectory commands and sends velocity commands to the STM32 microcontroller to control the motors.

\vspace{-0.7cm}

\subsection{Aerial Platform}
\label{aerial_platform}

\vspace{-0.5cm}

Our aerial platform is a self-developed quadrotor with four DJI Snail 2305 motors installed to serve as actuators. A DJI A3 flight controller is utilized as the autopilot to control the attitude of the drone. Meanwhile, the IMU inside together with the forward-looking MatrixVision mvBlueFox-MLC200w global shutter camera with a 235$^{\circ}$-FOV fish-eye lens performs state estimation to localize the aerial vehicle. A downward-looking Stereolabs ZED stereo camera with an FOV of $110^{\circ} \times 80^{\circ}$ and a range of 20 m is mounted below the vehicle to produce depth images for the mapping. The computations are all performed onboard on an Intel i7-5500U processor and an Nvidia Jetson TX2, both of which are running Ubuntu 16.04 and ROS Kinetic.

The software running onboard is responsible for localization, mapping and exploration planning. The VINS-Mono from our previous work~\cite{vins}, which fuses IMU data and image data from the forward-looking camera together, is adopted for localization of the aerial vehicle. Besides the control purpose, the fused odometries are also sent to the exploration planner for planning. The SDK of the ZED stereo camera provides depth images and point clouds for dense mapping and the point clouds are transformed and merged into the global map.

\vspace{-0.5cm}

\section{Experiments and Results}
\label{experiments}

\vspace{-0.2cm}

\subsection{Outdoor Autonomous Exploration}
\label{subsec:outdoor_experiment}

\vspace{-0.5cm}

The outdoor experiment site is an \(17m \times 8m\) area, with various obstacles, as shown in Fig.~\ref{fig:experiment_site}. 
in Fig ~\ref{fig:real_data}, the quadrotor and the ground vehicle both start from the bottom to perform exploration. The downward-looking camera is configured to have a \(2m \times 2m\) FOV. The UAV is flying at a constant height of 2 m with maximum speed 1.4 m/s, while the UGV is configured to have a laser scanned range of 6 m and travels at a maximum speed of 0.5 m/s. The exploration results are shown in Fig.~\ref{fig:real_data} and Fig.~\ref{fig:real_prog}.

\vspace{-0.6cm}

\subsection{Simulative Exploration}
\label{subsec:sim_experiment}

\vspace{-0.5cm}

We also perform simulations on distinctive maps. All environments are $20m\times20m$ in size and have a 0.1-m resolution. Exploration starts with both robots placed at a corner. The UGV travels at a maximum speed of 1m/s, while the UAV has a higher maximum speed of 1.4 m/s. The simulated UAV is equipped with a downward-looking depth camera with a FOV of $60^{\circ} \times 60^{\circ}$ in the vertical and horizontal directions, and the UGV carries a $360^{\circ}$ laser scanner with a 6-m range. The UAV hovers at a constant height of 2 m, resulting in a 4.6-m$^2$ sensor coverage. The exploration results are shown in Fig.~\ref{fig:sim_data} and Fig.~\ref{fig:maze_prog}.

\begin{figure}[t]
\centering
\includegraphics[width=0.95\columnwidth]{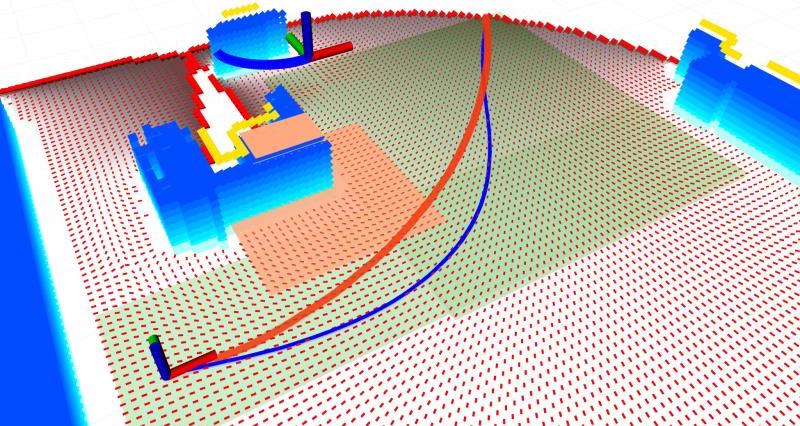}
\caption{Red and yellow points are Frontier A/B cells respectively. The orange point cloud is from the UAV's sensor. Darkness represents the harmonic function value, and red vectors represent the cells' gradient. The field guides the UGV to a further but longer frontier line. Green squares are the travel corridor, and the blue and red curve on the ground are the original gradient-descent and the optimized trajectory respectively. The blue curve above is the trajectory planned for the UAV.
\label{fig:sim_capture}}
\vspace{-0.5cm}
\end{figure}

\vspace{-0.7cm}

\subsection{Comparisons and Analyses}
\label{subsec:comparisons_analyses}

\vspace{-0.5cm}

\begin{figure}
%\captionsetup{justification=centering}
\subfigure[\label{fig:maze_traj} Collaborative trajectory]
{\includegraphics[width=0.48\columnwidth]{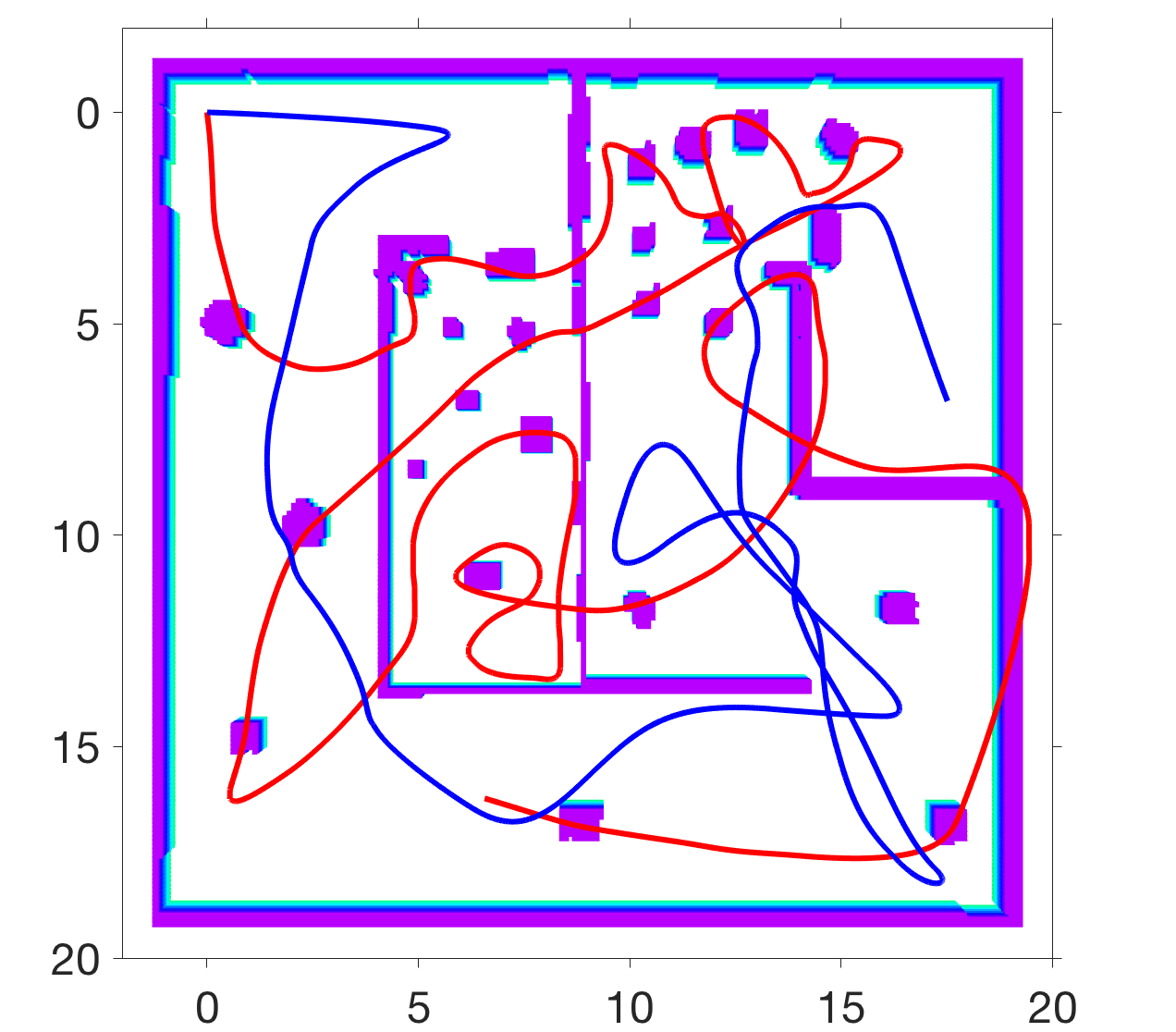}}
\subfigure[\label{fig:maze_single_traj} Individual trajectory]
{\includegraphics[width=0.48\columnwidth]{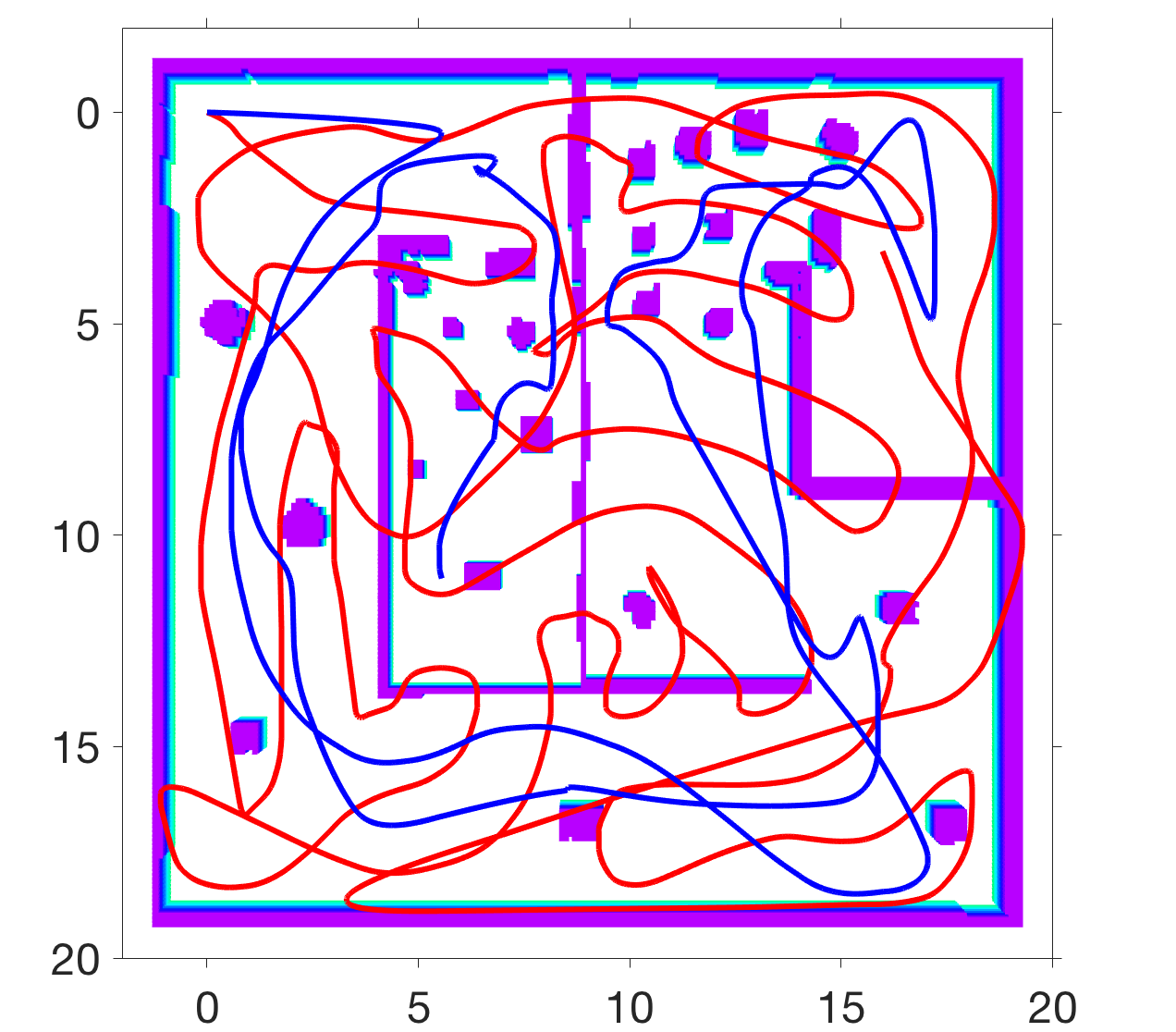}}
\caption{Exploration results of the maze map. Red curve represents UAV; blue curve represents UGV.\label{fig:sim_data}}
\end{figure}

\begin{figure}
%\vspace{-0.5cm}
%\captionsetup{justification=centering}
\centering
\subfigure[\label{fig:real_traj} Collaborative trajectory]
{\includegraphics[width=0.48\columnwidth]{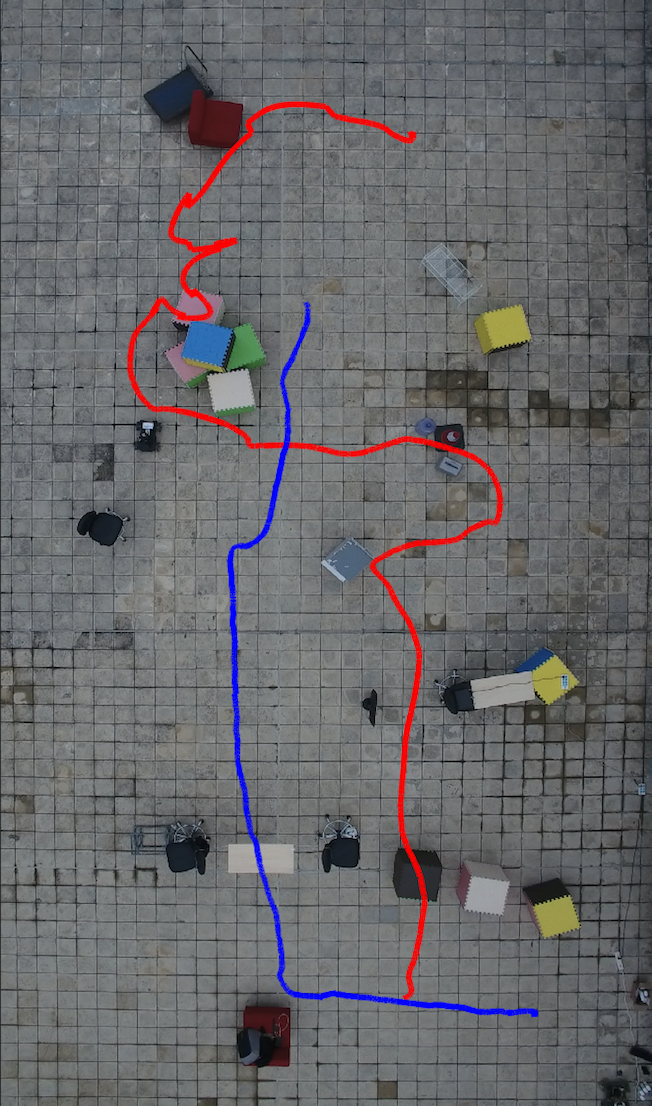}}
\subfigure[\label{fig:real_single_traj} Individual Trajectory]
{\includegraphics[width=0.48\columnwidth]{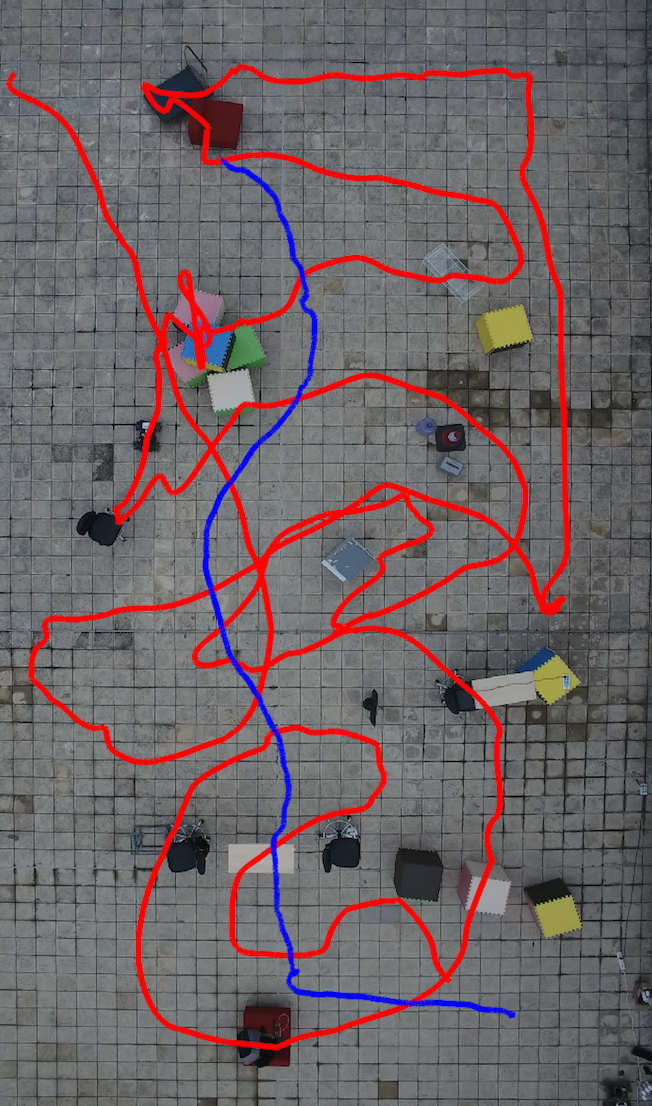}}
\caption{Exploration results of the experiment site. Red curve represents UAV; blue curve represents UGV.\label{fig:real_data}}
%\vspace{-0.5cm}
\end{figure}

\begin{figure}
\captionsetup{justification=centering}
\centering
\subfigure[\label{fig:maze_prog} Maze]
{\includegraphics[width=0.48\columnwidth]{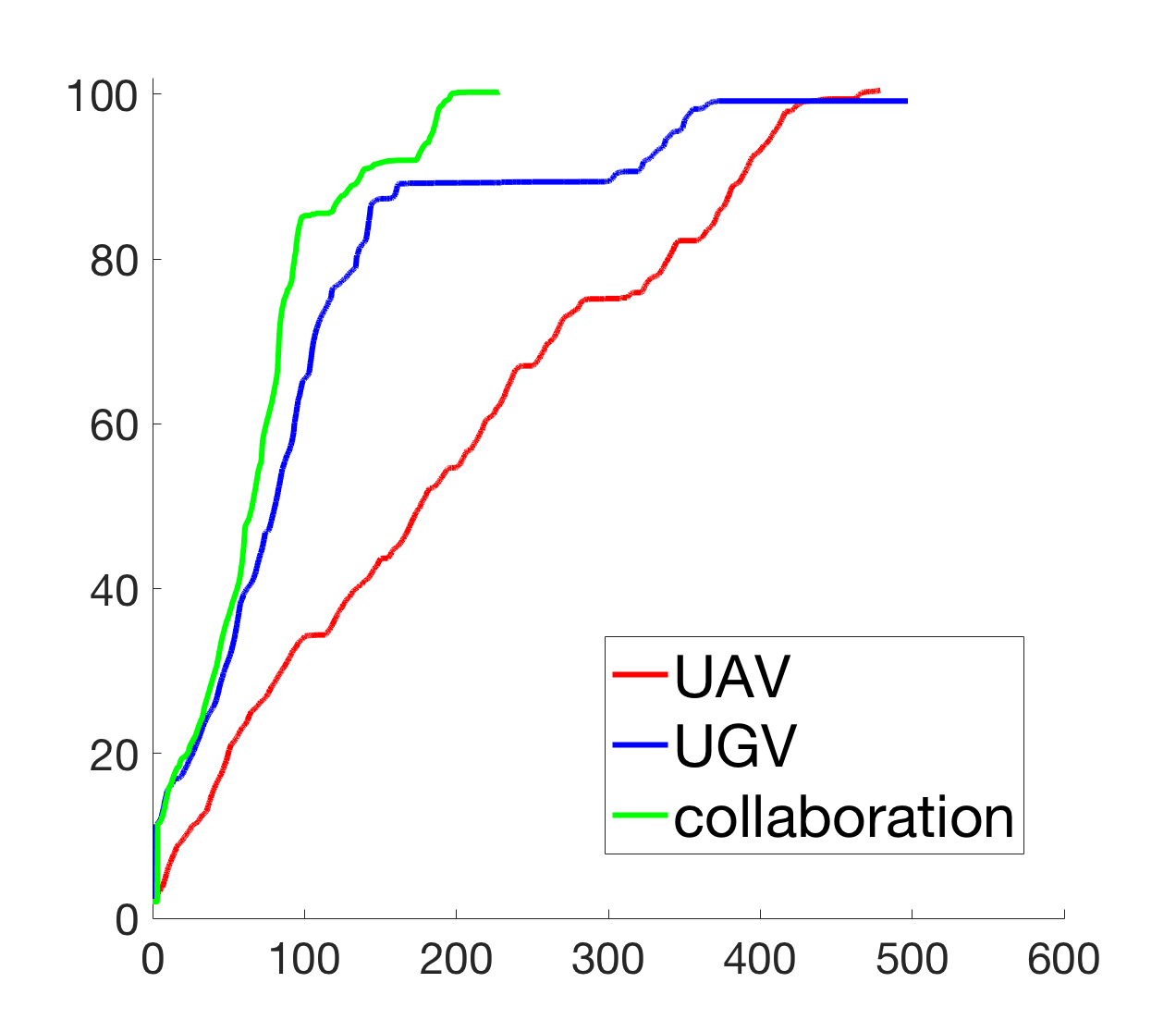}}
\subfigure[\label{fig:real_prog} Outdoor Experiment]
{\includegraphics[width=0.48\columnwidth]{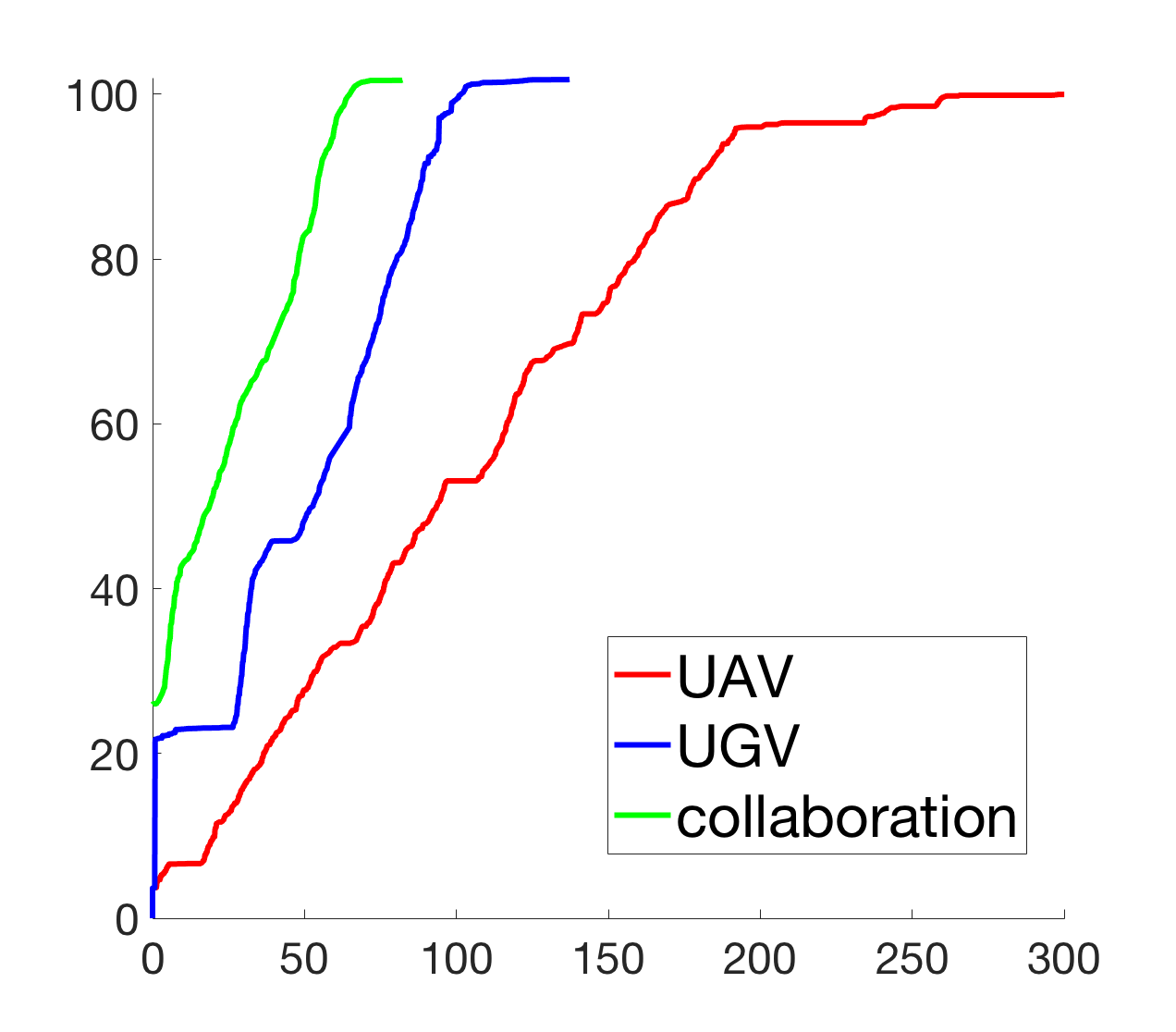}}
\caption{Exploration progress in simlation and outdoor experiment. \label{fig:prog}}
\end{figure}

Fig.~\ref{fig:sim_capture} shows a visualization-capture, as an illustration of the framework. To examine the framework, we compare the performance (progress against time) of the UAV-UGV team, a single UAV, and a single UGV. Fig.~\ref{fig:sim_data}, Fig.~\ref{fig:real_data} and Fig.~\ref{fig:prog} show the exploration performance in simulations and outdoor experiments. We also test the sequences mentioned in Sec.~\ref{subsec:UGV_planning}, and the results are shown in Table~\ref{tab:table_2}.

\begin{table}[t]
\begin{center}
\caption{Comparison of Exploration Time and Trajectory Generation Time}\label{tab:table_1}
\begin{tabular}{|c|c|c|c|c|c|c|}
\hline
\multirow{2}{*}{Environment Type} & \multirow{2}{*}{$T_{exp}^{UGV (1)}$ (s)} & \multirow{2}{*}{$T_{exp}^{UAV (1)}$(s)} & \multirow{2}{*}{$T_{exp}^{col (1)}$ (s)} & \multirow{2}{*}{$\eta_{col}^{(2)}$} & {UAV Avg. } & {UGV Avg.} \\ 
&&&&& {$t_{traj.gen.}^{(3)}$ (ms)} & {$t_{traj.gen. }$ (ms)}\\ \hline
Maze & 366.25 & 448.75 & \textbf{139.50} & 1.45 & 27.3 & 21.5\\ \hline
Fence and Clusters & 278.75 & 448.75 & \textbf{178.00} & 0.97 & 28.5 & 23.9\\ \hline
Uniform Random & 146.80 & 401.50 & \textbf{119.75} & 0.90 & 28.9 & 17.4 \\ \hline
Outdoor Experiment Site$^{(3)}$ & 94 & 191 & \textbf{59} & 1.07 & 73.9 & 45.0\\ \hline
\end{tabular}
\end{center}
(1) $T_{exp}^{UGV}$, $T_{exp}^{UAV}$ and $T_{exp}^{col}$ denote the time duration from the start of the exploration to exploration progress of 95\% of the UGV alone, the UAV alone and the collaborative system respectively, since the interior of the obstacles cannot be fully scanned by the UGV.

(2) $\eta_{col}$ denotes the collaborative efficiency, which is defined as:
\begin{equation}
\label{eq:col_efficiency}
\eta_{col}=\frac{\frac{1}{T_{exp}^{col}}}{\frac{1}{T_{exp}^{UGV}}+\frac{1}{T_{exp}^{UAV}}}
\end{equation}

(3) $t_{traj.gen.}$ denotes the computation time for trajectory generation.

(4) Note that the simulation and the outdoor experiment platforms are different. The trajectory generation time section of the outdoor experiment is just for reference.

\vspace{-0.5cm}
\end{table}

As shown in Fig.~\ref{fig:prog}, the collaborative team has significantly higher exploration efficiency than individual vehicles. The time for exploration of the system is significantly reduced, as is the trajectory length of each vehicle, as shown in Fig.~\ref{fig:sim_data} and Fig.~\ref{fig:real_data}. According to the results shown in Table \ref{tab:table_1}, the collaborative system has at least 23\% higher exploration speed compared with UGV alone, at least 145\% higher exploration speed compared with UAV alone and the advantages can reach up to 163\% and 235\% respectively. From the definition of the collaborative efficiency $\eta_{col}$ in Eq.~\ref{eq:col_efficiency}, the variable can reflect the effectiveness of the collaboration. For the worst of the testing cases, uniform random, the collaborative efficiency can still reach 0.90, even though the exploration site is the least suitable for utilizing the advantages of both vehicles. For the intuitively most favorable case, the maze, the efficiency can reach 1.45, indicating the proper modeling of the system. The high collaborative efficiency reflects the effectiveness of the work distribution of the vehicles. Meanwhile, the computation times of the trajectories can all be restricted to below 80 ms, meaning that the method can be adopted for real-time online navigation.

\begin{table}[t]
\begin{center}
\caption{Comparison of the Different Update Sequences of the Harmonic Field}\label{tab:table_2}
\begin{tabular}{|c|c|c|c|}
\hline
\multirow{2}{*}{Environment Type} & {Convergence Iterations of}&{Convergence Iterations of} & {Iteration Reduction of} \\ 
& {Conventional Sequence} & {BFS Sequence}&{BFS in Percentage} \\ \hline
Maze & 268.837 & \textbf{250.068} &  6.98\% \\ \hline
Fence and Clusters & 347.648 & \textbf{331.658} & 4.60\% \\ \hline
Uniform Random & 396.084 & \textbf{365.863} & 7.63\% \\ \hline
\end{tabular}
\end{center}
\vspace{-0.8cm}
\end{table}

In Table~\ref{tab:table_2}, the proposed BFS update sequence of the harmonic field mentioned in Sec.~\ref{subsec:UGV_planning} is compared with the normal row-by-row update sequence. It is clearly shown that the BFS sequence consumes fewer iterations for each of the cases, hence reducing the computation power by 4.60\% to 7.63\%.

\vspace{-0.5cm}

\section{Experimental Insights}

\vspace{-0.5cm}

Control is always the most basic issue in robotics experiment. The control of a quadrotor system can be affected by multiple factors. The error from the actuators can significantly deteriorate the performance, affecting other parts of the system. On our aerial platform, the original DJI 2312E motors with a maximum thrust of 850-g/ rotor was substituted with DJI 2305 Snail motors with a maximum thrust of 1.32-kg/ rotor. Reserving a larger margin on the actuators can reduce the uncertainty in the control segment, enhancing the control performance, and it can further increase the reliability in field applications.

In this experiment, the re-planning strategy is one of the most crucial issues. Time allocation is still a significant problem in path planning and it can produce serious effects on the consistency of the generated trajectories. In our system, to avoid the inconsistency caused by improper time allocation, we take trajectory length, time, and initial and end state into consideration to produce an effective re-planning strategy.

\vspace{-0.5cm}

\section{Conclusion}

\vspace{-0.5cm}

In this paper, we propose an autonomous aerial-ground collaborative exploration system. The system consists of a UGV and a UAV and utilizes their complementary characteristics to achieve high efficiency in the exploration process. The ground vehicle carries a laser scanner to localize itself with the IMU and construct the map while traveling amid obstacles; the UAV utilizes a forward-looking camera to localize itself with the installed IMU, while constructing the map through a downward-looking stereo camera when flying above obstacles. The proposed exploration planning method combines a frontier method, a field method and a motion primitive method to plan efficient and consistent motions for the vehicles. 

%For the ground vehicle, employing the frontier and obstacle information, a harmonic field is constructed to obtain a gradient-descent path to the frontier. Then, the path is utilized to generate a safe corridor, where a piece-wise Bernstein polynomial basis trajectory is optimized against the squared jerk and the generated trajectory is sent to the vehicle to execute. For the aerial vehicle, a dense motion primitive tree is firstly generated in the surrounding area with limited velocity. Then, according to the score of the primitives, which is formulated by the information gain and the cost, the primitive possessing the highest score is selected. Afterwards, the path to the primitive is utilized to create a safe corridor where the piece-wise Bernstein polynomial trajectory is optimized against the squared jerk. In the case that no primitive has positive information gain, a frontier cell with the highest score is chosen and the final trajectory is optimized using minimum jerk trajectory generator. 
Extensive simulated experiments are performed to validate the proposed method. Furthermore, the aerial and ground platforms, which are integrated with state estimation, dense mapping and exploration planning modules, are implemented to verify the practicability of the method in an outdoor environment. The results show that the collaborative system has superior exploration speed against each vehicle alone and has a high collaborative efficiency.

\vspace{-0.7cm}

\bibliographystyle{plain}
\bibliography{iser2018luqi}
\end{document}